\documentclass[pdflatex,sn-mathphys]{sn-jnl}

\jyear{2023}%

\theoremstyle{thmstyleone}%
\theoremstyle{thmstyletwo}%

\theoremstyle{thmstylethree}%

\usepackage{subfigure}
\usepackage{multirow}

\newlength{\bibitemsep}
\newlength{\bibparskip}
\let\oldthebibliography\thebibliography
\renewcommand\thebibliography[1]{%
  \oldthebibliography{#1}%
  \setlength{\parskip}{\bibitemsep}%
  \setlength{\itemsep}{\bibparskip}%
}

\begin{document}

\title[Transformers for the Detection of Cerebral Aneurysms in Microsurgery]{Shifted-Windows Transformers for the Detection of Cerebral Aneurysms in Microsurgery}


\author*[1,2]{\fnm{Jinfan} \sur{Zhou}}\email{\{jinfan.zhou}
\equalcont{These authors contributed equally to this work.}

\author[1]{\fnm{William} \sur{Muirhead}}\email{\{w.muirhead}
\equalcont{These authors contributed equally to this work.}

\author[1]{\fnm{Simon C.} \sur{Williams}}\email{s.c.williams}

\author[1]{\fnm{Danail} \sur{Stoyanov}}\email{danail.stoyanov} 

\author[1]{\fnm{Hani J.} \sur{Marcus}}\email{h.marcus\}@ucl.ac.uk} 

\author*[1]{\fnm{Evangelos B.} \sur{Mazomenos} }\email{e.mazomenos\}@ucl.ac.uk} 

\affil*[1]{\orgdiv{Wellcome/EPSRC Centre for Interventional and Surgical Sciences}, \orgname{Universtiy College London}, \orgaddress{ \city{London}, \postcode{WC1E 6BT}, \country{UK}}}

\affil[2]{\orgdiv{Robotics Institute}, \orgname{University of Michigan}, \city{Ann Arbor}, \state{MI}, \postcode{48109}, \country{USA}}


\abstract{\textbf{Purpose} Microsurgical Aneurysm Clipping Surgery (MACS) carries a high risk for intraoperative aneurysm rupture. Automated recognition of instances when the aneurysm is exposed in the surgical video would be a valuable reference point for neuronavigation, indicating phase transitioning and more importantly designating moments of high risk for rupture. This article introduces the MACS dataset containing 16 surgical videos with frame-level expert annotations and proposes a learning methodology for surgical scene understanding identifying video frames with the aneurysm present in the operating microscope's field-of-view.   \\
\textbf{Methods} Despite the dataset imbalance (80\% no presence, 20\% presence) and developed without explicit annotations, we demonstrate the applicability of Transformer-based deep learning architectures (MACSSwin-T, vidMACSSwin-T) to detect the aneurysm and classify MACS frames accordingly. We evaluate the proposed models in multiple-fold cross-validation experiments with independent sets and in an unseen set of 15 images against 10 human experts (neurosurgeons).  \\
\textbf{Results} Average (across folds) accuracy of 80.8\% (range 78.5\%-82.4\%) and 87.1\% (range 85.1\%-91.3\%) is obtained for the image- and video-level approach respectively, demonstrating that the models effectively learn the classification task. Qualitative evaluation of the models’ class activation maps show these to be localized on the aneurysm's actual location. Depending on the decision threshold, MACSWin-T achieves 66.7\% to 86.7\% accuracy in the unseen images, compared to 82\% of human raters, with moderate to strong correlation.  \\
\textbf{Conclusions} Proposed architectures show robust performance and with an adjusted threshold promoting detection of the underrepresented (aneurysm presence) class, comparable to human expert accuracy. Our work represents the first step towards landmark detection in MACS with the aim to inform surgical teams to attend to high-risk moments, taking precautionary measures to avoid rupturing.}


\maketitle
\section{Introduction}\label{sec1}
Neurosurgery is heavily reliant both on preoperative imaging (CT/MRI) and microscopy for intraoperative visualisation. Neuronavigation using frameless stereotaxy is widely used in neurosurgical procedures, with considerable interest into integrating augmented reality (AR) for injecting an overlay of anatomically useful information into the optics of a microscope, tracked by the stereotaxy system \cite{lee2020vision, cabrilo14}. As the navigation system is unable to interpret the anatomical information contained in the surgical scene itself, both systems rely on manual registration of surface (patient’s head) landmarks. However, neuronavigation becomes inaccurate as the brain moves intraoperatively \cite{kantelhardt15}. This limits the utility of AR overlays for fine microsurgery such as neurovascular procedures~\cite{meola2017augmented}. Automated recognition of anatomical landmarks in Microsurgical Aneurysm Clipping Surgery (MACS), and broadly within neurosurgery is particularly relevant as it has the potential to recalibrate stereotactic registration. Detection of critical anatomy enhances interpretation of the surgical environment and can bridge the gap between preoperative imaging and intraoperative surgical video~\cite{chadebecq2020computervi}. Artificial Intelligence (AI)-based detection of relevant anatomy has been previously explored in abdominal, laparoscopic and neck surgery~\cite{madani2021artificial, tokyasu2020dev, gong2021using}.

This study focuses on the automated detection of the aneurysm itself. This target is interesting for multiple reasons. Firstly, because the part of the surgery when the aneurysm is in view is recognized as being particularly high risk of aneurysm rupture ~\cite{muirhead2021adverse}. A system which can alert the wider theatre team that they are entering a high-risk phase of the procedure could facilitate a coordinated response should rupture occurs. Secondly, because the aneurysm and its associated vessels are typically the focus of the overlay in AR-supported MACS, automated detection and localization may determine which AR overlays would be of maximum value to the surgeon. As the most relevant anatomical target in MACS, it can also provide a reference landmark for recalibration of stereotaxy. Finally, the exposure of the aneurysm in the field-of-view, typically marks a phase transition from cisternal dissection into aneurysm neck dissection. Recognition of the aneurysm could therefore contribute to automated operative workflow analysis~\cite{khan2021automated}. Further innovations can also focus on surgical education and operative planning/decision making.

We collected and present a microsurgical aneurysm clipping surgery (MACS) dataset comprising of 16 videos of MACS procedures with aneurysm presence/ absence annotations at frame-level ($\sim$350k frames) conducted by expert neurosurgeons\textbf{.} The aneurysms' size and appearance in the video varies significantly among cases, also depending on surgical approach, while its overall visual appearance is similar to adjacent brain vasculature. Both these aspects pose interesting challenges for vision-based classification in the absence of precise annotations (i.e. bounding boxes) and with class labels only providing weak supervision.

\begin{figure*}[t!]
\centering
\subfigure{
\includegraphics[width=12cm]{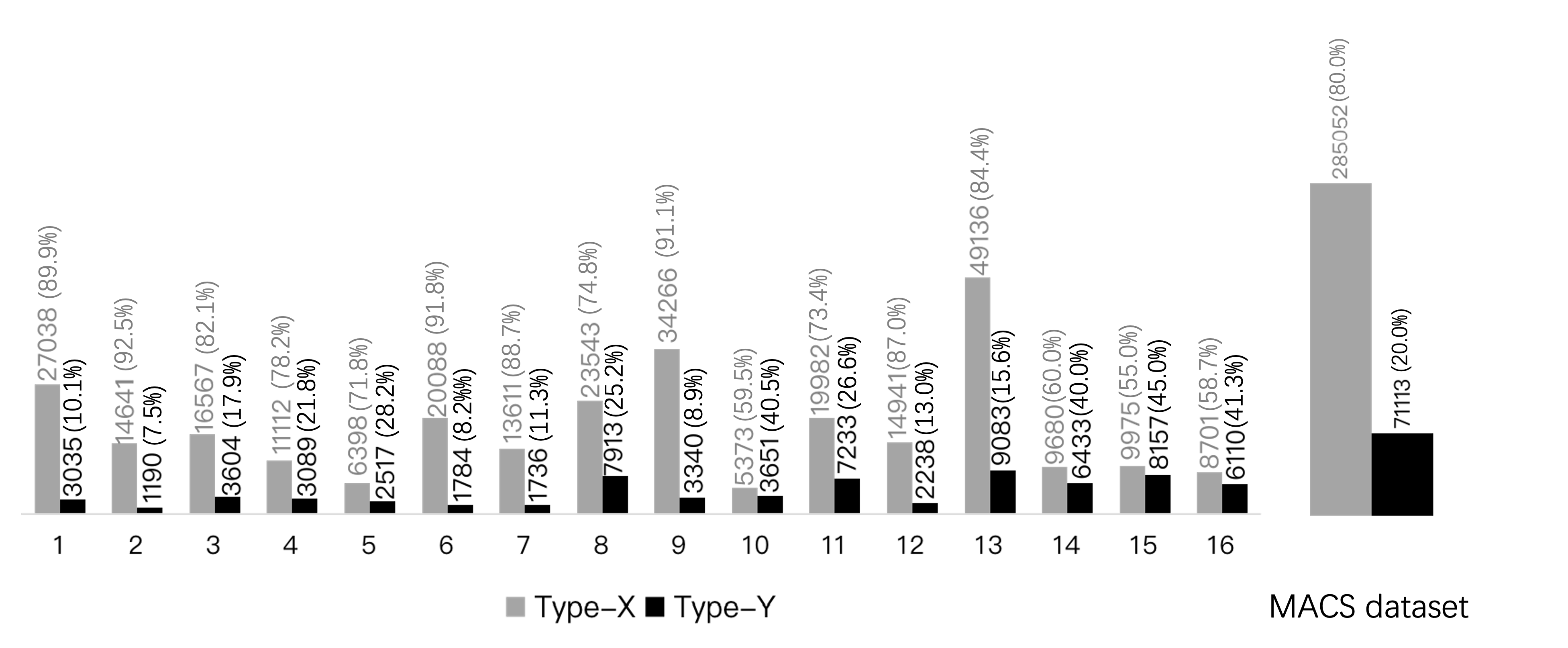}
}
\caption{Dataset distribution of the MACS dataset. Each video comprises two types of frames for analysis, namely \textit{Type-X} and \textit{Type-Y}. }
\label{fig1}
\end{figure*}

We develop and evaluate two deep learning architectures, an image-based (MACSSwin-T) and a video-based (vidMACSSwin-T) for performing aneurysm detection and classification, based on the lightweight version of the shifted-windows Transformer model (Swin-T)~\cite{liu2021swin}. Attention-based learning architectures have been previously adapted for surgical video analysis and applied in tasks such as depth estimation~\cite{long2021dssr}, phase recognition~\cite{czempiel2021opera} and instruction generation~\cite{zhang2021surgical}.

Incorporating hierarchical, multi-scale self-attention \cite{vaswani2017attention, dosovitskiy2020image}, the proposed Swin-T model extracts localized, representations at different levels enabling the network to learn features for detecting and distinguishing the aneurysm. Our base model (MACSSwin-T) is a frame-level classification Swin-T architecture, which is expanded to a video-based model (vidMACSSwin-T), incorporating temporal information by aggregating features from multiple successive frames. In multiple-fold cross-validation experiments with independent sets covering the entire available dataset, MACSSwin-T and vidMACSSwin-T achieve 80.8\% and 87.1\% average classification accuracy respectively, demonstrating the efficiency of our approach. We further compare the performance of MACSSwin-T against human evaluators (10 London-based consultant neurosurgeons) on an external set of 15 microscopy images, with the MACSSwin-T having similar performance (13/15 - 86.7\%) to human experts (12.3/15 - 82\%), when lowering the decision threshold, without compromising false positives.

\section{Methods}\label{sec2}
\begin{figure}[t!]
\centering
\subfigure[]{
\includegraphics[width=4.6cm]{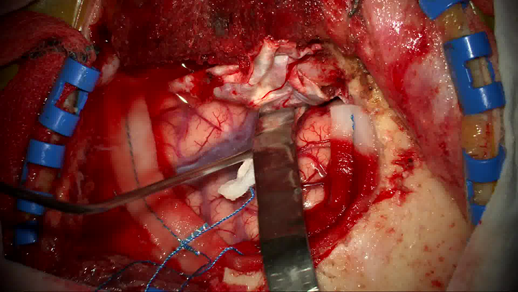}
\label{fig2a}
}
\quad
\subfigure[]{
\includegraphics[width=4.6cm]{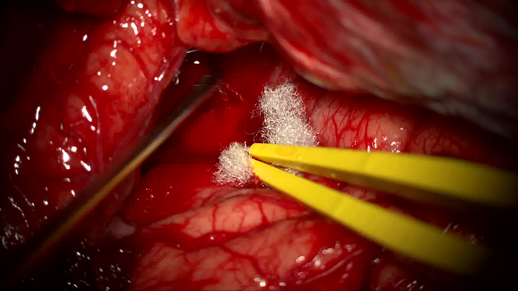}
\label{fig2b}
}
\quad
\subfigure[]{
\includegraphics[width=4.6cm]{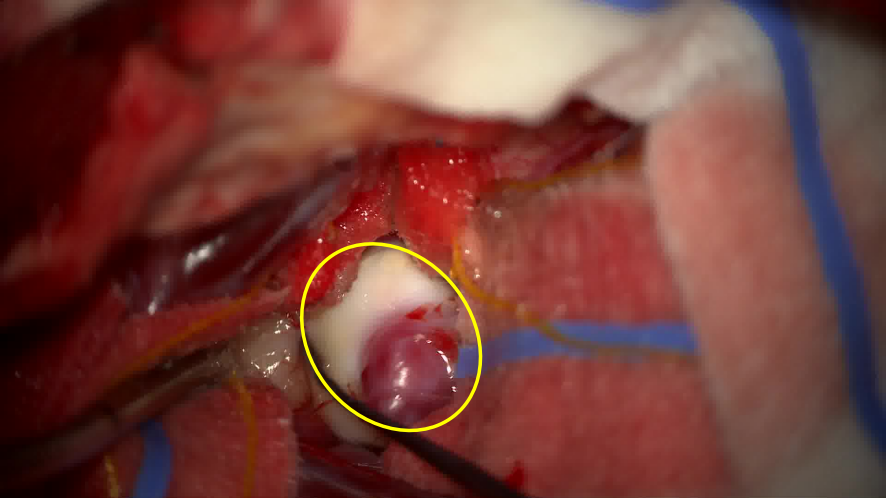}
\label{fig2c}
}
\quad
\subfigure[]{
\includegraphics[width=4.6cm]{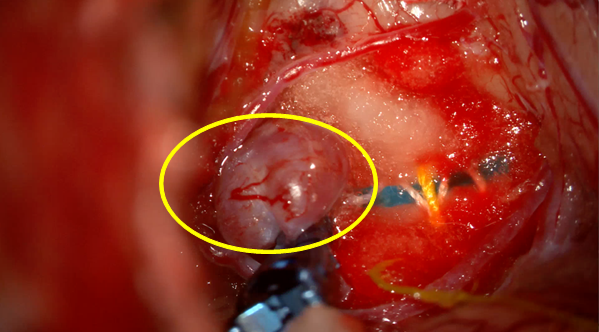}
\label{fig2d}
}
\quad
\subfigure[]{
\includegraphics[width=4.6cm]{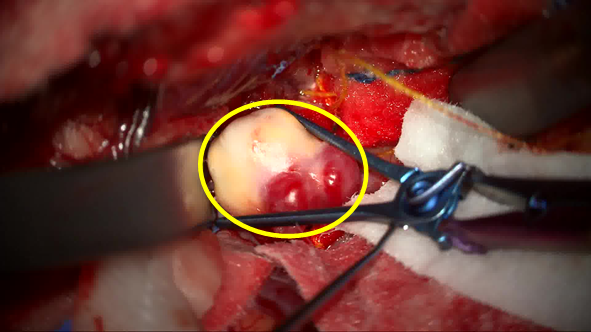}
\label{fig2e}
}
\quad
\subfigure[]{
\includegraphics[width=4.6cm]{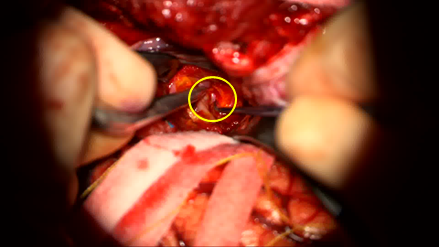}
\label{fig2f}
}
\quad
\subfigure[]{
\includegraphics[width=4.6cm]{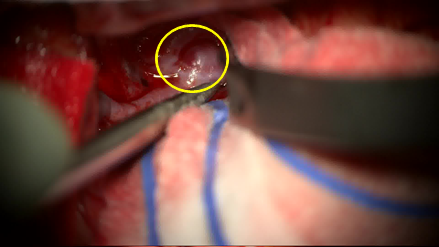}
\label{fig2g}
}
\quad
\subfigure[]{
\includegraphics[width=4.6cm]{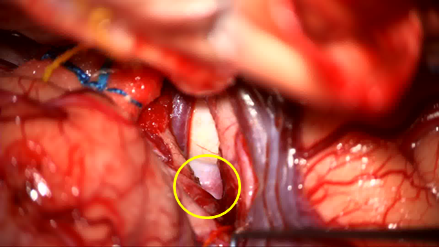}
\label{fig2h}
}
\caption{Excerpts from the MACS dataset. Fig.~\ref{fig2a} and Fig.~\ref{fig2b} shows \textit{Type-X} examples. Fig.~\ref{fig2c}-Fig.~\ref{fig2h} are \textit{Type-Y} examples with the yellow oval indicating the location of the aneurysm (Fig.~\ref{fig2d} shows a clipped aneurysm). Fig.~\ref{fig2f}-Fig.~\ref{fig2h} are examples of challenging \textit{Type-Y} frames where only a small part of the aneurysm is visible.}
\label{fig2}
\end{figure}
\subsection{The Microsurgical Aneurysm Clipping Surgery (MACS) Dataset}
The MACS dataset is composed of FHD (1920x1080) surgical videos from the operative microscope of 16 patients during surgical repair of intracerebral aneurysms. The study was registered with the hospital’s (National Hospital for Neurology and Neurosurgery, UCLH NHS) local audit committee and data sharing was approved by the information governance lead. All patients provided written informed consent for their videos to be collected for research. The MACS dataset was blindly reviewed by two senior vascular neurosurgeons in duplicate. Frames were classified as follows: \textit{Type-X}: No aneurysm in microscope's view, \textit{Type-Y}: Aneurysm in microscope's view (including both visible and clipped aneurysms), \textit{Type-Z}: Frame excluded from analysis due to one or more from the following \textbf{i)} microscope not pointing at patient, \textbf{ii)} microscope moving, \textbf{iii)} indocyanine green angiography being run, \textbf{iv)} ambiguous image with partial view of the aneurysm making it inconclusive to assign either \textit{Type-X} or \textit{Type-Y} label, \textbf{v)} instruments crossing the field-of-view resulting in the aneurysm rapidly entering and exiting the field-of-view, \textbf{vi)} rapid changing view within the scene. Frames with conflicting labels from the senior reviewers were also excluded from the dataset (Type-Z). Whilst it can be difficult to identify an aneurysm from reviewing just a single frame, over the course of the entire video each aneurysm was necessarily clearly identified to make clipping possible. Annotating surgeons had not only been present at the time of surgery but also had the entire video available to contextualise each frame. Examples of \textit{Type-X} and \textit{Type-Y}, labelled images are illustrated in Fig.~\ref{fig2}.

We extract frames at 5 fps and establish a dataset of 356,165 images. Frames labelled as \textit{Type-Z} are discarded, keeping only \textit{Type-X} and \textit{Type-Y}. The distribution of labeled images per video recording is shown in Fig.~\ref{fig1}. The dataset presents high imbalance between the \textit{Type-X} and \textit{Type-Y} classes. This is expected since in practice during the MACS procedure the aneurysm remains within the microscope's field-of-view for a limited amount of time typically when aneurysm dissection is performed. The imbalanced distribution of the two classes in the MACS dataset, poses an interesting challenge on developing learning methods to detect short-duration but critical events (i.e. aneurysm in the field-of-view) in image-guided surgical procedures.

\subsection{Proposed learning architectures}
Primary objective of our work is to develop a methodology to automatically detect the presence of the aneurysm in microscopy video frames, without any explicit location information(i.e. bounding-boxes) about a visible aneurysm or surgical tools (incl. clips) being available. The task presents challenges due to three key reasons: \textbf{i)} the short duration of the aneurysm appearing in the microscope's field-of-view, resulting in an imbalanced distribution of \textit{Type-X} and \textit{Type-Y} frames, \textbf{ii)} intraclass difference between aneurysms leading to limited common features and \textbf{iii)} the variable, and in most cases, small size and similar visual appearance (color and morphological texture) of the aneurysm, compared to the rest of the brain vasculature which is also present in the surgical scene.

\begin{figure*}[ht]
\centering
\includegraphics[width=10cm]{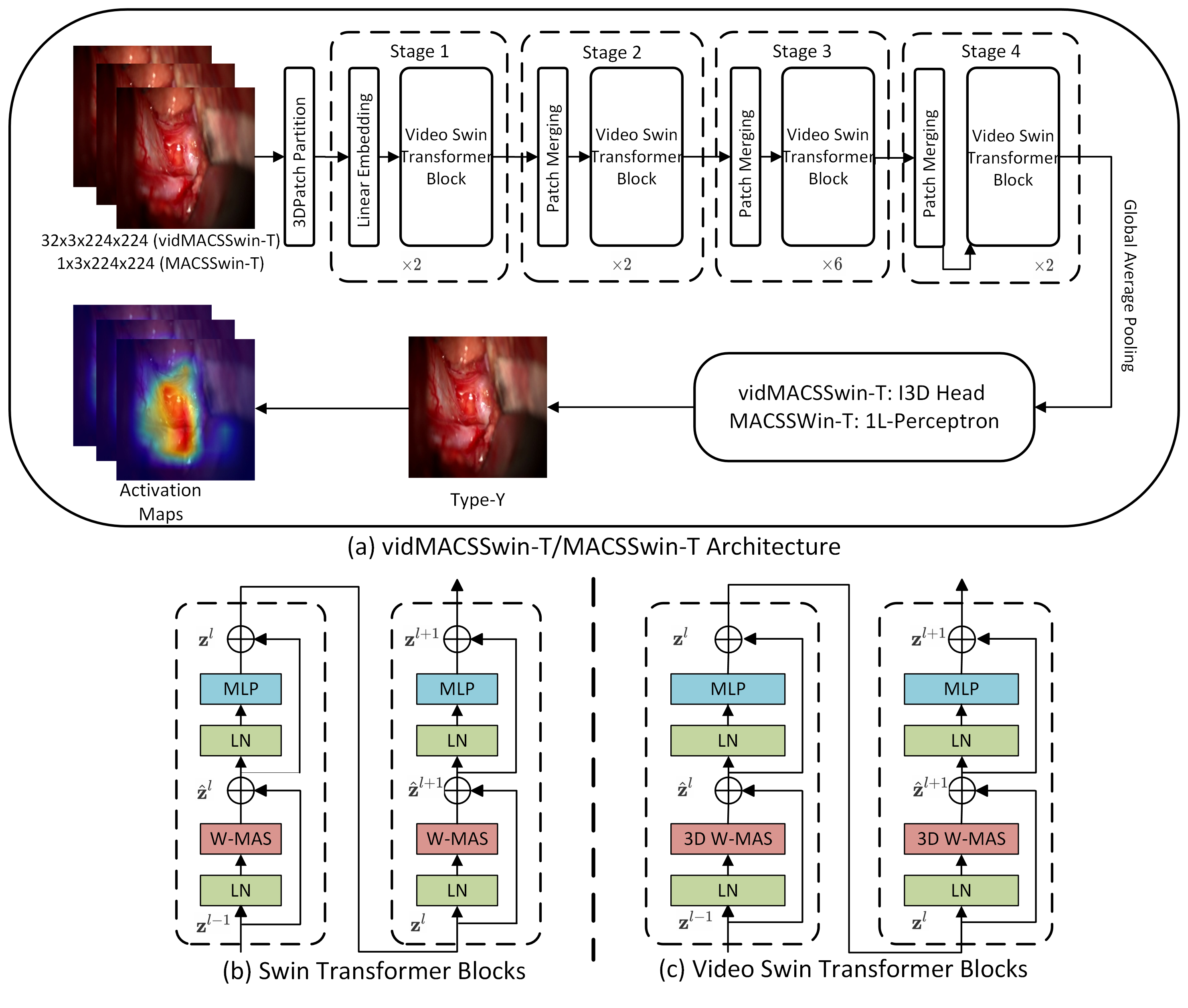}
\caption{(a): The proposed vidMACSSwin-T architecture for aneurysm detection. Features extracted from the 4 stages are fed to the aneurysm classification head, to produce the final prediction (i.e. \textit{Type-X} or \textit{Type-Y}); (b) and (c): detailed structures for Swin Transformer blocks and Video Swin Transformer Blocks.}
\label{fig3}
\end{figure*}

We formulate our problem as a frame classification task and adapt the tiny version of the Transformer model using shifted-windows (Swin-T)~\cite{liu2021swin} to tackle it. The proposed architecture is illustrated in Fig.\ref{fig3}. The MACSSwin-T model extracts features at 4 stages, where each stage consists of multiple consecutive Swin Transformer blocks. Each block is composed by a shifted-window multi-head self attention (MSA) layer and a 2-layer MLP with GELU activation functions in between. Global average pooling is applied to the feature maps, resulting in a 768-dimensional feature vector, processed by a single-layer perceptron with softmax activation to predict the final class (aneurysm presence/absence).

We incorporate temporal frame sequences to formulate aneurysm detection as a video classification task and propose vidMACSSwin-T (shown also in Fig.\ref{fig3}), adapted from~\cite{liu2021video}, to address it. The vidMACSSwin-T model is a 3D version of the MACSSwin-T model, which takes multiple frames (32) as input. It maintains the general structure of MACSSwin-T, while expanding the token to be a 3D patch. Accordingly, we replace the standard MSA in the Swin-T block with the 3D shifted window MSA module and keep the rest of the components unchanged. We use the I3D head~\cite{carreira2017quo} to obtain the output and use it as the prediction of the center frame.

Weighted cross entropy is the loss function to account for the data imbalance in the dataset. In Eq~\ref{eq:wcbe}, $\hat{y_i}$ and $y_i$ are the predicted score and ground-truth label respectively of sample $i$, and $N_b$ is the number of batch samples. We set the weights ($w_X$, $w_Y$) for each class as the ratio of the number of samples of the other class ($N_Y$, $N_X$) over the total samples and normalize them to add to "1". $N_X$ and $N_Y$ is the total number of \textit{type-X} and \textit{type-Y} samples in MACS.
\vspace{-10pt}
\begin{align} \label{eq:wcbe}
    L_{wbce} &= {\frac{1}{N_b}}\sum_{i=1}^{N_b}\left[w_Y y_i\log \hat{y_i} + w_X (1-y_i)\log (1-\hat{y_i}) \right] \\
    w_X &= \frac{N_Y}{N_X + N_Y}, w_Y = \frac{N_X}{N_X + N_Y} \nonumber
\end{align}
\section{Experiments and results}

\subsection{Implementation details}
We initialize MACSSwin-T with a pretrained version of Swin-T on ImageNet-21K. We then freeze the first 3 stages and train the network, for 300 epochs, on our MACS dataset using a batch size of 128. Input images are normalized and resized to 224x224, due to memory and timing considerations. We follow the data augmentation methods in~\cite{liu2021swin}. AdamW is employed for optimization with a base learning rate of 5e-4 and exponential decay rates $\beta_1$ and $\beta_2$ of 0.9 and 0.99. We warm up the training for 20 epochs with the warm up learning rate to be 5e-7. VidMACSSwin-T is initialized with a pretrained Swin-T model on ImageNet-21K~\cite{liu2021video}. Hyperparameters were set according to~\cite{liu2021video, liu2021swin}, and experimentation showed that their value has small effect on the performance of the MACSSwin-T/vidMACSSwin-T models. Frames are resized to 224$\times$224 without cropping. We train the model with AdamW optimizer for 30 epochs, using initial learning rates of 3e-5 and 3e-4, for the backbone network and I3D prediction head, respectively. We use a cosine decay learning rate scheduler and 3 epochs of linear warm-up. The batch size is 64. Because the MACS videos in our dataset are long (10k-50k frames), we divide each video into multiple samples. Following a similar approach to~\cite{liu2021video}, for training vidMACSSwin-T, we group 64 consecutive frames of the same class as one sample and from each sample, uniformly extract a sequence of 32 frames as the input to the network. During inference, for all extracted frames, we group $64\times4$ consecutive frames to define a single sample. We then uniformly sample these ($64\times4$) consecutive frames to formulate 4 sequences of 32 frames, perform predictions on all 4 sequences and average them to produce the final classification label. Models and experimentation took place with the PyTorch (v1.12.1) framework. We develop the models on an NVIDIA RTX A6000 GPU. Training MACSSwin-T requires about 6417MB of memory and inference time is 10.67 msec (97.3 fps).

\subsection{Multiple-fold cross-validation}
\begin{figure*}[!h]
\centering
\subfigure{
\includegraphics[width=11cm]{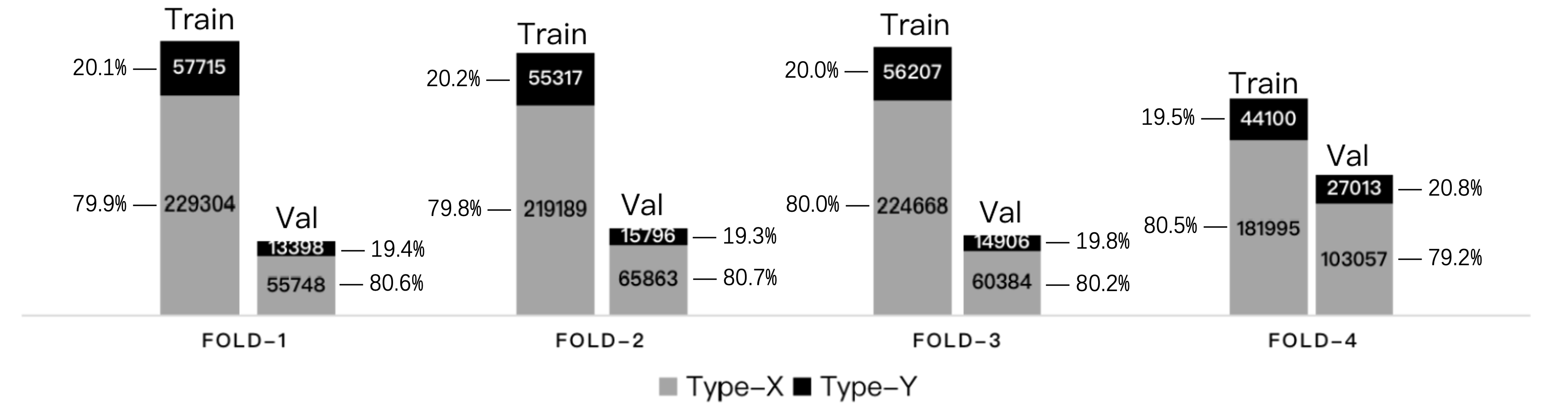}
}
\caption{Number of training and validation samples for each cross-validation fold.}
\label{fig4}
\end{figure*}

We carry out multiple-fold cross-validation and split the dataset into 4 folds on the basis of the 16 available videos creating independent training (12 videos) and validation (4 videos) sets in each fold. This partitioning allows us to evaluate the model's performance and consistency on the entire MACS dataset. The training/validation splits are selected to have a similar class ratio (ranging between 3.82:1 to 4.17:1) for both the training and validation sets, shown in Fig.~\ref{fig4}. Loss weights are set to $W_X = 0.2$ and $W_Y = 0.8$, in all folds according to the overall MACS dataset distribution.

\begin{table}\centering
\caption{Results from 4-fold cross-validation experiments}\label{table::results}
\begin{tabular}{|c|c|c|c|c|c|}
\hline
\textbf{Model} & Fold &  Accuracy(\%) & Precision(\%) & Recall(\%) & F1 score(\%)\\
\hline
\multirow{2}{8em}{MACSSwin-T} & 1, 2 &  81.3, 78.5 & 51.3, 45.5 &  66.5, 55.3 & 58.0, 49.9 \\
& 3, 4 & 80.9, 82.4 & 51.5, 56.9 & 70.0, 63.4 & 59.4, 60.0 \\
\hline
\textbf{Mean}  & - & \textbf{80.8}& \textbf{51.3}&
\textbf{63.8} &\textbf{56.8} \\
\hline
\multirow{2}{8em}{vidMACSSwin-T} & 1, 2 &  86.2,  85.1& 65.1, 66.4 &  61.0, 42.3 & 63.0, 51.7 \\
& 3, 4 & 91.3, 85.8 & 93.0, 93.0 & 59.5, 32.8 & 72.5, 48.4 \\
\hline
\textbf{Mean} & - & \textbf{87.1}& \textbf{79.4}& \textbf{48.9} &\textbf{58.9} \\
\hline
\end{tabular}
\end{table}

Results are listed in Table~\ref{table::results} for both models in each fold. For the single-frame MACSSwin-T model, the validation accuracy ranges from 78.5\% to 82.4\% with a mean accuracy of 80.8\%. The mean precision and recall rates are 51.3\% and 63.8\% respectively, resulting in a mean F1-score of 56.8\%. The results are promising indicating that the MACSSwin-T architecture learns to correctly recognize the presence/absence of the aneurysm. The model avoids introducing significant bias towards the dominant \textit{type-X} class, since the recall rate is significantly higher than the precision rate. Experiments with different initialisation seeds demonstrate robust model behaviour (std$<0.1$ on all 4 metrics).

The vidMACSSwin-T model using temporal information achieves a significantly higher mean accuracy at 87.1\% compared to MACSSwin-T. Average precision increases to 79.4\%, while the recall is 48.9\%. Across folds, the F1 score of vidMACSSwin-T is higher than that of MACSSwin-T. We argue that the result (85.8\%) of the fourth fold is not due to the model being biased to negatives. Although the recall is low (32.8\%), the proportion of negatives is 79.2\%, while the accuracy is higher (85.8\% - exceeding the highest accuracy 82.4\% of MACSSwin-T), which means vidMACSSwin-T still has the ability to recognize positives samples. Overall, the cross-fold validation experiments with independent sets, justify the selection of the Swin-T architecture and proposed models development strategy for the intended aneurysm classification task.

\subsection {Visualizing the Class Activation Maps}
\begin{figure*}[ht]
\centering
\subfigure{
\includegraphics[width=11.5cm]{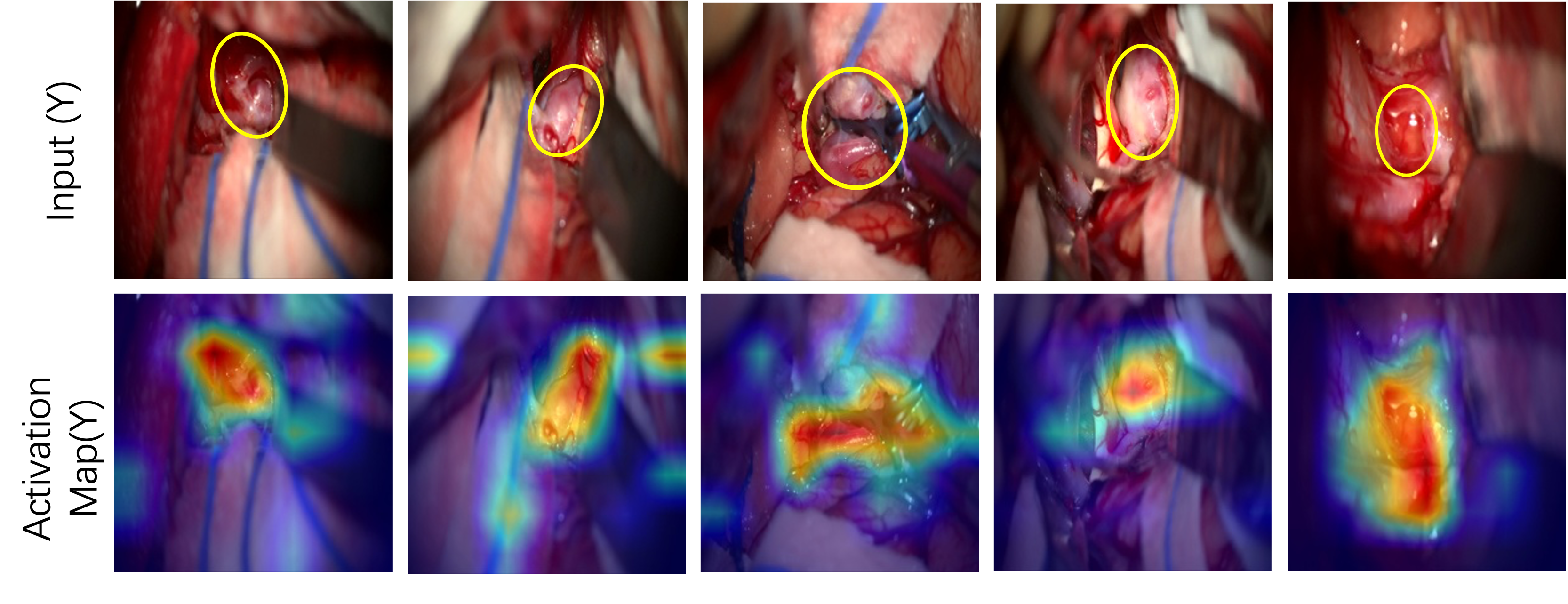}
}
\caption{Activation maps of the Swin-T model from \textit{Type-Y} input images. The first row shows the aneurysm areas as annotated by the expert neurosurgeons. The second row shows the activation maps of the final normalization layer after training is completed. The model's activation is localized in the same areas as the manual annotations.}
\label{fig5}
\end{figure*}

Fig.\ref{fig5} illustrates the class activation maps for 5 \textit{type-Y} input frames, using  Grad-CAM~\cite{selvaraju2017grad}, obtained at the final normalization layer after the completion of training in each fold. By comparing neurosurgeons' annotations (provided independently for these 5 input images) of the aneurysm location (top row) to the generated activation maps (bottom row), we conclude that the self-attention mechanism on the MACSSwin-T, drives the model to focus on the desired location of the image and correctly localizes the exposed aneurysms. We also highlight that this is achieved in the presence of adjacent vasculature with similar appearance and without providing to the network, any information on the aneurysms' location.

\subsection{Comparison against humans (neurosurgeons)}
To demonstrate the performance of our method against a gold standard and establish a preliminary baseline, 10 consultant neurosurgeons, 2 females and 8 males (age 35-64) from London-based NHS trusts, were surveyed. Human assessors were asked to classify, after visual inspection, 15 frames (8 Type-X and 7 Type-Y) selected from microscopy videos from 4 MACS procedures not included in the 16 ones, we used for model development. Specifically, frames were extracted from the 4 operative videos, unseen to the models at a rate of 5 fps. Fifty frames (50) were initially selected randomly and underwent blinded senior review by two vascular neurosurgeons in duplicate, where frames were classified as Type-X (aneurysm not in-frame), Type-Y (aneurysm in-frame), or Type-Z (excluded). The final dataset of 15 images shared to the 10 expert neurosurgeons was randomly selected from images with concordant reviews from the pool of 50 images. Human assessors reviewed and labelled the 15 test images as still frames, without access to the videos of the procedures. A total of 150 individual frame-level reviews (80 Type-X: no aneurysm, 70 type-Y: aneurysm present) was obtained. Human classifications were tested against the outputs of the MACSSwin-T model, that makes an inference based on a single frame. For completeness we report results  with the vidMACSSwin-T, but do not perform a direct comparison as the human assessors reviewed individual images, instead of videos. Models were retrained on the entire MACS dataset and tested only on the 15 images.

\begin{figure*}[t!]
\centering
\subfigure[]{
\includegraphics[width=4.6cm]{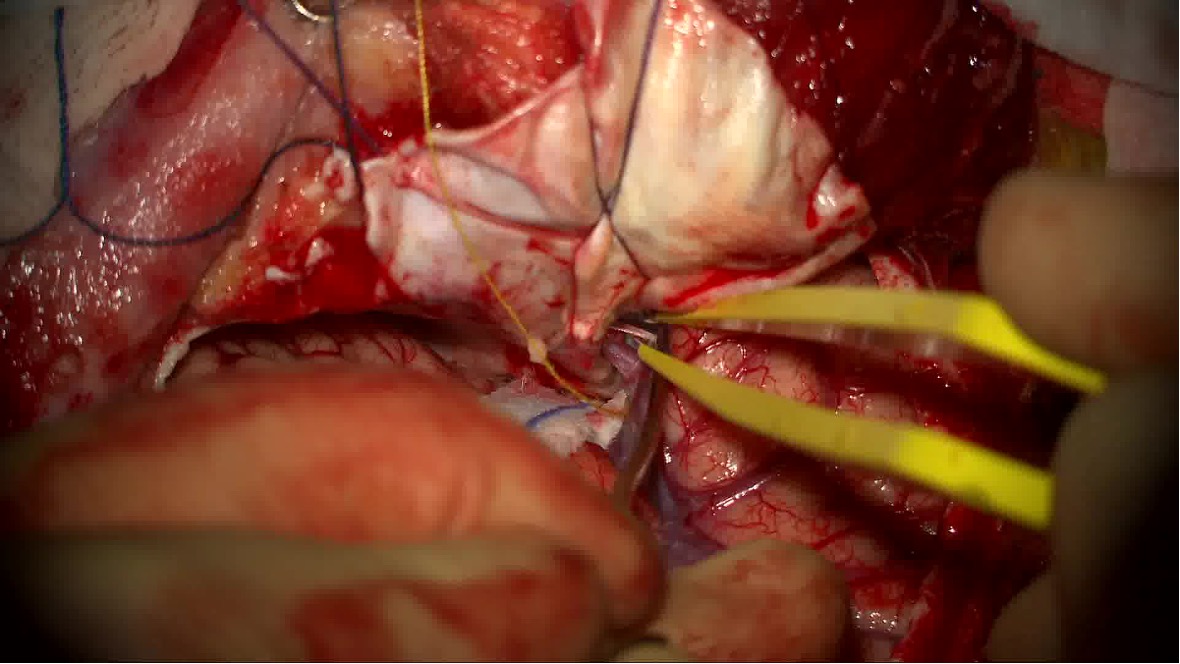} 
\label{fig6a}
}
\quad
\subfigure[]{
\includegraphics[width=4.6cm]{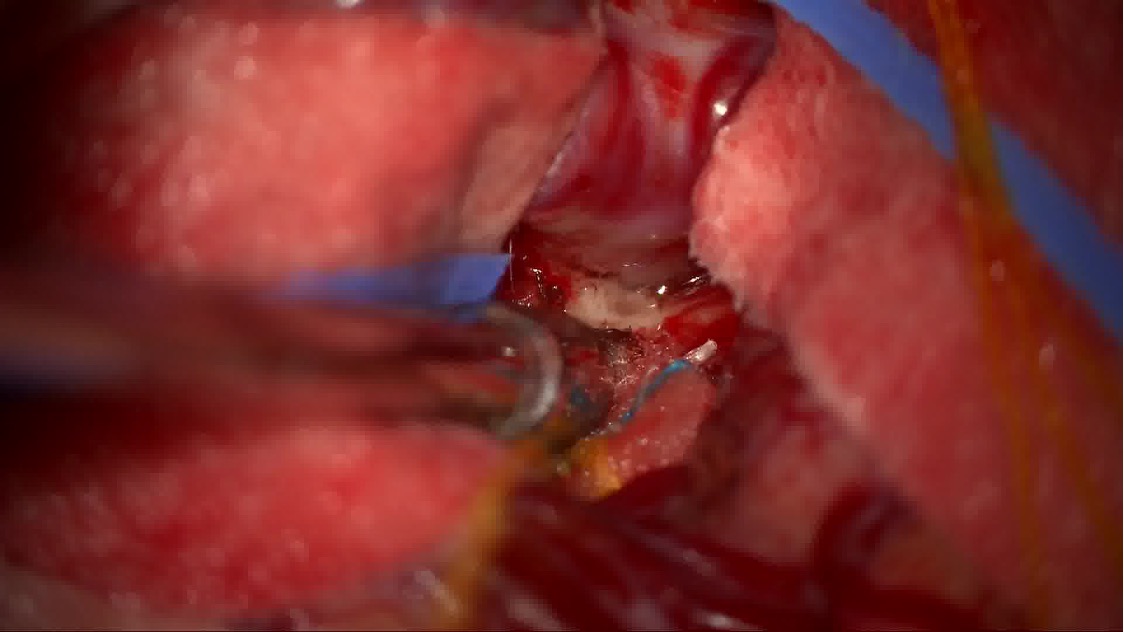} 
\label{fig6b}
}
\quad
\subfigure[]{
\includegraphics[width=4.6cm]{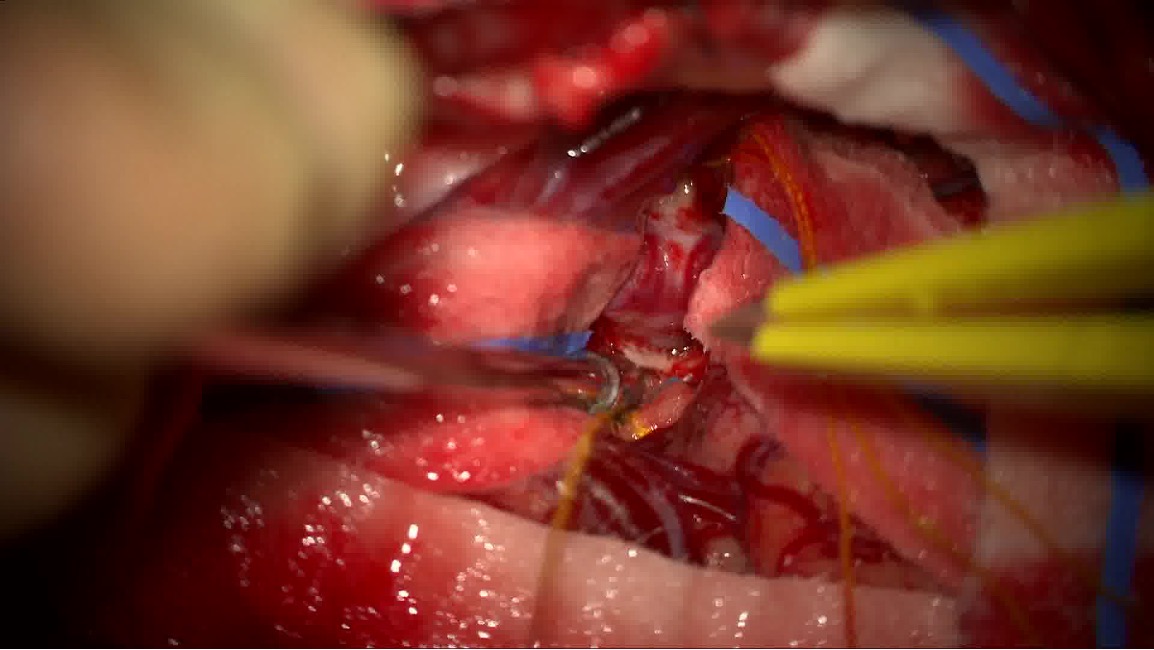} 
\label{fig6c}
}
\quad
\subfigure[]{
\includegraphics[width=4.6cm]{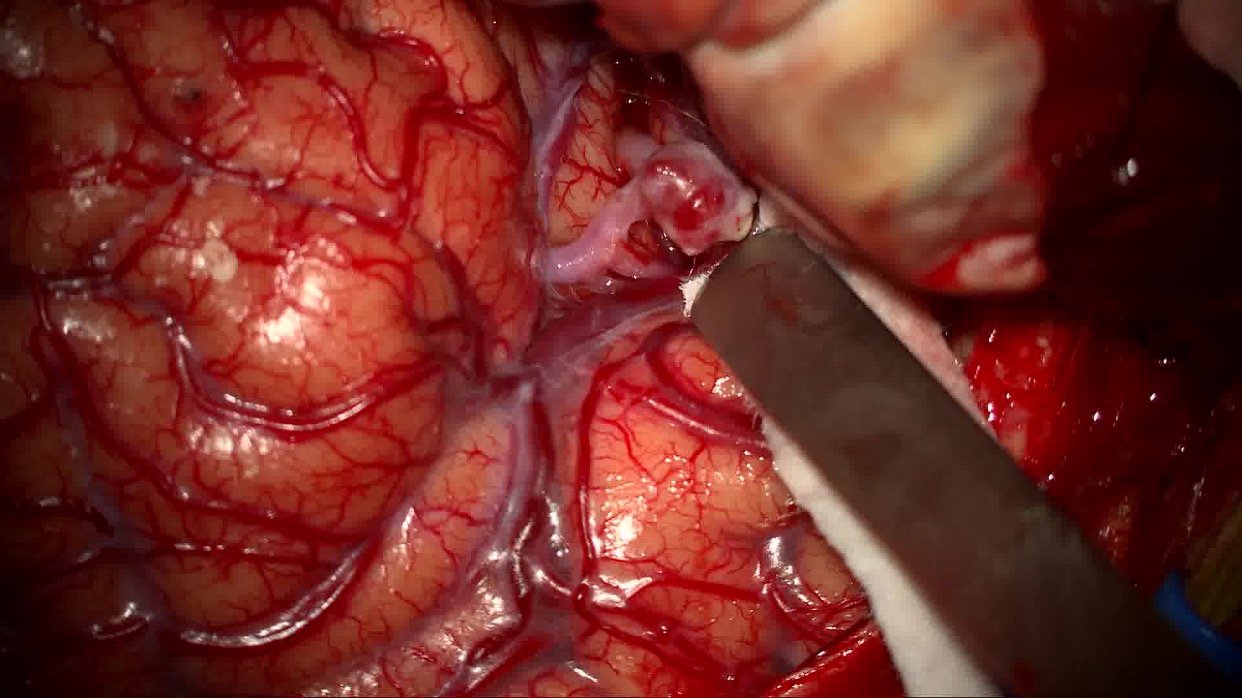} 
\label{fig6d}
}
\quad
\subfigure[]{
\includegraphics[width=4.6cm]{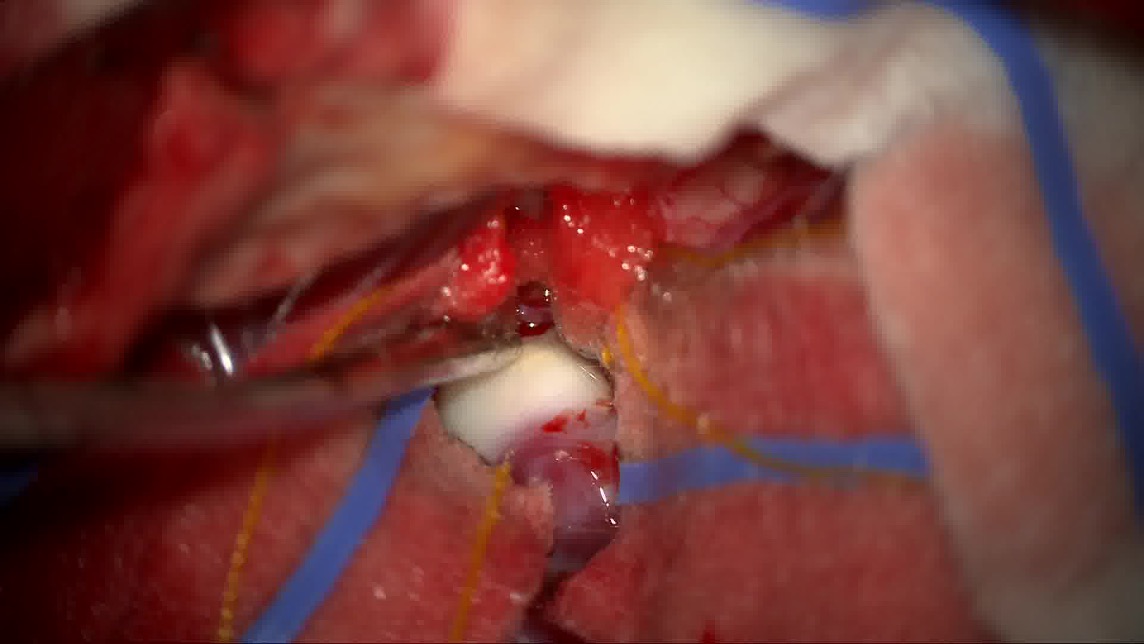} 
\label{fig6e}
}
\quad
\subfigure[]{
\includegraphics[width=4.6cm]{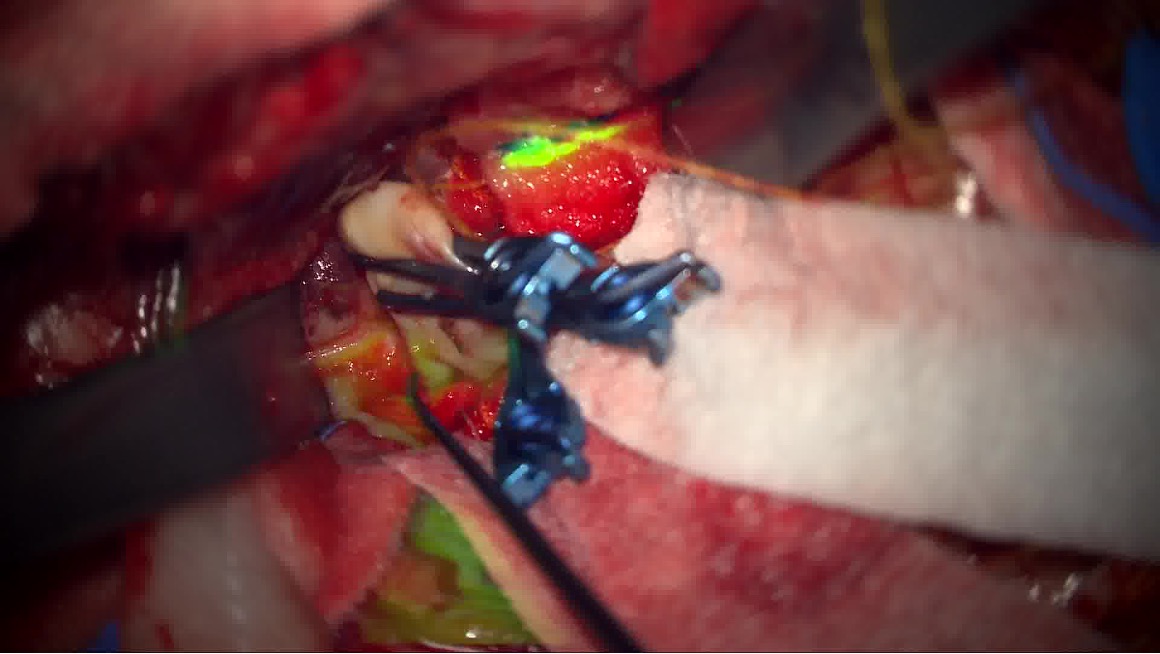} 
\label{fig6f}
}
\caption{Challenging image samples (from the set of 15) yielding detection errors for both neurosurgeons and AI (MACSSWin-T) models. Fig.~\ref{fig6a}, is negative (Type-X) however, the vessel the bipolar forceps points to, is probably misinterpreted by humans as the aneurysm leading to low detection rate (40\%). AI consistently detects it as Type-X. Fig.~\ref{fig6b}-\ref{fig6d} are positive (Type-Y), but only a tiny part of the aneurysms is visible and also they visually blend (overlay) over adjacent vessels which possibly makes it difficult to distinguish. Humans achieve (50\%, 50\% and 100\%) while AI initially fails with $DT=0.5$, but correctly classifies Fig.~\ref{fig6b},~\ref{fig6d} with $DT=0.3$. Videos of atherosclerosed (white dome) aneurysms like in Fig.~\ref{fig6e} and with indocyanine green like in Fig.~\ref{fig6f} are rarely seen during training. Due to the largely different visual appearance, AI initially ($DT = 0.5$) fails to recognize these two aneurysms. Setting $DT = 0.4$, gives correct classification for Fig~\ref{fig6f} showing that the clip (blue color), typically an indication of exposure, provides a strong enough visual cue for AI to correctly identify the aneurysm. Humans assisted by the presence of the white dome and clip achieve very good performance in these two cases (90\%, 100\%). Fig.~\ref{fig6c},~\ref{fig6e} are the ones AI fails to detect correctly.}

\label{fig6}
\end{figure*}

Neurosurgeon and model results are listed in Table \ref{table::human-ai comparison}.
In total, 123/150 (82\%), of human reviews were correct in identifying or excluding the presence of an aneurysm. For Type-X images,  67/80 (84\%) reviews correctly excluded the presence of an aneurysm, while for Type-Y frames,  56/70 (80\%) reviews correctly identified the presence of an aneurysm in the image. Neurosurgeons' individual accuracy ranged from 68.7\%(10/15) to 100\%(15/15) with 11/15 (73.3\%) and 13/15(86.7\%) observed in most cases (3 each). Due to the significant class imbalance in the MACS dataset, we expect our models to be biased towards Type-X (negative) samples. In order to promote prediction of Type-Y (positive) samples, we reduce the decision threshold ($DT$) of the output softmax layer, in MACSSwin-T from 0.5 (initial) to 0.3, and investigate the model's behaviour when encouraged to classify Type-Y samples with lower confidence. We have experimentally verified in CV experiments that smaller than 0.3 threshold values do not further promote the detection of Type-Y samples. MACSSwin-T accuracy ranges between 66.7\% an 86.7\%, with excellent precision (100\%). We observe that the recall rate improves with reduced thresholds (0.3, 0.4) and the model correctly recognizes up to 3 Type-Y samples previously classified as Type-X. We also remark that none of the Type-X images is wrongly classified with decreased $DT$. Overall, lowering the $DT$ is beneficial for the model as it correctly classifies previously missed Type-Y samples albeit with lower confidence. Different to our model that always (for all $DTs$) classifies Type-X frames correctly, humans find this challenging and sometimes misdetect them, probably because of tool presence or adjacent vasculature being perceived as the aneurysm. Fig.~\ref{fig6} shows challenging samples where misclassifications are observed from both humans and the MACSSWin-T model.

Inter-rater agreement, between the 10 human assessors, was found to be moderate with Fleiss Kappa value $0.502$ ($p$-value = $0$, agreement is not accidental). Strong agreement is seen in 8 samples (5 negatives and 3 positives) with 9 or more similar classifications, while 1 negative (4neg-6pos, Fig~\ref{fig6a}) and 2 positives (5neg-5pos, Fig~\ref{fig6b} and ~\ref{fig6c}) have very weak agreement. The agreement between the MACSSwin-T model and humans is evaluated with the Matthews correlation coefficient ($MCC$), which for binary classification tasks equals the Pearson correlation. We provide the range of $MCC$ for MACSSwin-T (for $DT=0.3$ and $DT=0.5$) and the 10 assessors. For $DT=0.5$ the $MCC$ ranges $0.08-0.48$ and shows moderate ($0.3-0.5$) positive correlation with the majority (8) of raters. For $DT=0.3$, the $MCC$ ranges $0-0.75$, showing strong ($0.5-0.7$) and very strong ($>0.7$), positive correlation with 7 raters. There is only one rater (who achieves 10/15 accuracy) with low correlation. This rater also has very low MCC ($<0.3$) with the remaining 9. Overall, MACSSwin-T presents moderate correlation against human raters for $DT=0.5$ and strong for $DT=0.3$.

\begin{table}\centering
\caption{Comparison against consultant neurosurgeons}\label{table::human-ai comparison}
\begin{tabular}{|c|c|c|c|c|}
\hline
 Method&  Accuracy(\%) & Precision(\%) & Recall(\%) & F1 score(\%)\\
\hline
MACSSwin-T(0.5) &  66.7(10/15) & \textbf{100.0(2/2)} & 28.6(2/7) & 44.4 \\
MACSSwin-T(0.4) &  73.3(11/15)& \textbf{100.0(3/3)} & 42.9(3/7) & 60.0\\
MACSSwin-T(0.3) &  \textbf{86.7(13/15)}& \textbf{100.0(5/5)} & 71.4(5/7) &\textbf{83.3}\\
vidMACSSwin-T  &  73.3(11/15)& \textbf{100.0(3/3 )} & 42.9(3/7) & 60.0\\
\hline
Human & 82.0(123/150 ) & 81.2(56/69) & \textbf{80.0(56/70)} & 80.6 \\
\hline
\end{tabular}
\end{table}

\section{Conclusion}
This article introduced a dataset of 16 MACS videos, with frame-level annotations on aneurysm presence/absence, and proposed Transformer-based models for automated frame-level aneurysm detection. In addition to having a small size, aneurysm exposure in MACS is a critical but short-term event occurring only during a particular procedural phase (aneurysm dissection). This results in the aneurysm annotated as visible, only in a fraction (ratio approximately 1:4 in our dataset) of video frames.

We develop our models (MACSSwin-T,  vidMACSSwin-T ), with frame-level annotations and weakly supervise them using cross-entropy loss with weights adjusted for the class imbalance in the MACS dataset. The self-attention module produces meaningful localized representations even in the absence of localized training signals (i.e. bounding boxes), enabling the models to efficiently detect the aneurysm and distinguish it from adjacent brain vasculature with similar appearance. We achieve consistent results with an average accuracy of 80.8\% (precision 51.3\%, recall 63.8\%) and 87.1\% (precision 79.4\%, recall 48.9\%), for the MACSSwin-T and vidMACSSwin-T models respectively in multiple-fold cross-validation, with independent training/validation sets. Although the task challenges are reflected in the obtained accuracy, our models achieve similar results to human assessors (86.7\% - 82\%), in an independent test set, with an adjusted detection threshold.

Future work will focus on three directions, i) pre-processing, ii) temporal information aggregation and iii) weakly supervision, to optimise model development.

\section*{Declarations}
\begin{unenumerate}
\item \textbf{Funding} This research was funded in whole, or in part, by the Wellcome/EPSRC Centre for Interventional and Surgical Sciences [203145/Z/16/Z]; the Engineering and Physical Sciences Research Council (EPSRC) [EP/P027938/1, EP/R004080/1, EP/P012841/1]; the Royal Academy of Engineering Chair in Emerging Technologies Scheme and the NIHR UCLH/UCL BRC Neuroscience. For the purpose of open access, the author has applied a CC BY public copyright licence to any author accepted manuscript version arising from this submission.
\item \textbf{Conflict of interest} The authors declare no conflict of interest.
\item \textbf{Ethics approval} This article does not contain any studies with human participants performed by any of the authors.
\item \textbf{Informed consent} This article does not contain patient data. Human assessors consented to anonymously participate in the survey presented.
\item \textbf{Data, code and/or material availability} Relevant code, models and the MACS dataset will be publicly released upon acceptance at the UCL WEISS Open Data Server https://www.ucl.ac.uk/interventional-surgical-sciences/weiss-open-research/weiss-open-data-server. Supplementary videos are attached, showing the MACSSWin-T model's predictions from the multiple-fold cross-validation experiments.
\end{unenumerate}

\bibliography{citation.bib}

\end{document}